\documentclass[11pt]{article}

\usepackage[final]{acl}

\usepackage{times}
\usepackage{latexsym}
\usepackage[T1]{fontenc}
\usepackage[utf8]{inputenc}
\usepackage{microtype}
\usepackage{float}
\usepackage{tabularx}
\usepackage{inconsolata}
\usepackage{bbm}

\usepackage{booktabs}

\usepackage{array}

\usepackage{amsmath,amssymb,amsfonts}
\usepackage{algorithmic}
\usepackage{graphicx}

\usepackage[table]{xcolor}
\usepackage{url}
\usepackage{makecell}

\usepackage{mdframed}
\usepackage[inline]{enumitem}
\usepackage{xspace}

\definecolor{graybg}{HTML}{F5F5F5}

\newcounter{prompt}
\newcommand{\boxedname}{Prompt\xspace}

\newenvironment{prompt}[1][]{
    \refstepcounter{prompt}
    \begin{mdframed}[
        innertopmargin=4pt,
        innerbottommargin=4pt,
        innerleftmargin=5pt,
        innerrightmargin=5pt,
        skipabove=6pt,
        skipbelow=6pt,
        linewidth=0.5pt,
        frametitle={\textbf{\boxedname\theprompt:} #1},
        frametitlefont=\small,
        frametitleaboveskip=3pt,
        frametitlebelowskip=3pt,
        frametitlerule=true,
        frametitlerulewidth=0.5pt,
        backgroundcolor=graybg
    ]
    \small
    \ttfamily
    \raggedright
    \setlength{\parindent}{0pt}
    \setlength{\parskip}{3pt}
    \setlength{\emergencystretch}{2em}
    \setlist[itemize]{
        leftmargin=*,
        topsep=2pt,
        itemsep=1pt,
        parsep=0pt,
        partopsep=0pt
    }
    \setlist[enumerate]{
        leftmargin=*,
        topsep=2pt,
        itemsep=1pt,
        parsep=0pt,
        partopsep=0pt
    }
}{
    \par
    \end{mdframed}
}

\newcommand{\diagtrace}{\textsc{DiagTrace}\xspace}

\title{DiagTrace: An Interactive Benchmark for Clinical Decision Making}

\author{
Yufan Wang$^{1}$ \quad
Rui Yang$^{1,2}$ \quad
Yi Lin$^{1}$ \quad
Chunhua Weng$^{3}$ \quad
Yifan Peng$^{1,2}$ \\
\\
$^{1}$Weill Cornell Medicine, New York, NY, USA \\
$^{2}$Cornell University, Ithaca, NY, USA \\
$^{3}$Columbia University, New York, NY, USA
}

\begin{document}
\maketitle
\begin{abstract}
Large language models (LLMs) have shown strong performance on medical question answering and diagnostic reasoning benchmarks. However, existing evaluations typically provide all patient information upfront and assess only the final diagnosis or rationale. This setup ignores the need to acquire diagnostic evidence incrementally from incomplete information in real-world diagnosis. To bridge this gap, we introduce \diagtrace, a source-grounded benchmark for evaluating whether LLMs can reconstruct the documented diagnostic trajectory through interactive information seeking. Each benchmark instance is built from a PubMed Central case report and includes an initial presentation, an event bank, a clinical course, and diagnostic references. \diagtrace contains 510 cases across eight specialties, with 225 used to evaluate four LLMs on clinical-course alignment, interaction quality, and efficiency. Results reveal trade-offs in coverage, source-consistent ordering, and turn use. 
\end{abstract}

\section{Introduction}
Recent medical large language models (LLMs) have primarily been evaluated in static settings~\citep{singhal2023clinicalknowledge,qiu2025medrbench,wang2024direct}, where a fixed set of patient findings is provided and LLMs predict a diagnosis or rationale. While useful for measuring medical knowledge and reasoning, these benchmarks do not test how a model actively acquires clinically relevant information.

This gap is \emph{important} because clinical diagnosis is not only an inference problem, but also an information-acquisition process, including deciding what to ask, what to examine, what to test, and when the accumulated findings are sufficient. 
Recent interactive medical benchmarks move closer to this setting by requiring models to ask follow-up questions or request tests over multiple turns~\citep{li2024mediq, tu2025conversational, liu2025interactive, fan2025aihospital, sviridov2025_3mdbench, chen2026clindiag, werthaim2026q4dx}. However, they typically prioritize final diagnostic accuracy or dialogue quality, leaving it under-evaluated whether a model can actively seek the right information at the right time. 

In this work, we ask: \emph{given the correct diagnosis, can a model actively reconstruct the diagnostic trajectory through interactive information seeking?} This formulation matters for two reasons.
\emph{First}, by providing the final diagnosis, it separates information acquisition from diagnostic inference. 
\emph{Second}, it supports evaluation under a fixed turn budget, requiring the model to prioritize informative requests and balance information coverage against interaction cost~\citep{yue2026interactivebenchmarks}.

We introduce \diagtrace, a benchmark built from PubMed Central (PMC) case reports for evaluating interactive diagnostic reconstruction \citep{zhao2023pmcpatients, zhang2025casereportbench, nlm2026pmcoa}. Each case includes an initial patient presentation, an event bank, a clinical course, and final diagnostic references. During evaluation, a clinician (evaluated) agent receives the initial presentation and final diagnosis, and must interact with a source-grounded report agent under a fixed turn budget to recover the supporting diagnostic trajectory. 
We evaluate this process at the trajectory level by comparing the model’s request trajectory with a source-derived reference using metrics like clinical course alignment, interactive quality, and turn efficiency and availability.

\diagtrace contains 510 cases spanning eight clinical specialties, with 225 cases used for comprehensive evaluation. 
Our results reveal substantial variation in models' ability to acquire diagnostic evidence, highlighting trade-offs between information coverage, source-consistent ordering, and interaction efficiency.

Our contributions are threefold:
\begin{enumerate*}[label=(\roman*)]
    \item a source-grounded benchmark derived from case reports;
    \item a clinical interactive task that isolates information acquisition from diagnosis inference; and
    \item a trajectory-level evaluation protocol for measuring what information models request, in what order, and at what interaction cost.
\end{enumerate*}

\section{Related Work}

Medical LLM evaluation has traditionally focused on static question answering and structured diagnostic reasoning benchmarks~\citep{singhal2023clinicalknowledge, wang2024direct, qiu2025medrbench, zhou2026medthink, choi2026ascent,yang2026clinmmbench}. While these benchmarks assess clinical knowledge and reasoning quality, the relevant findings are provided rather than actively acquired. In contrast, \diagtrace evaluates how clinician agents actively reconstruct a diagnostic trajectory through sequential information seeking.

Interactive clinical benchmarks move closer to real diagnostic settings by evaluating models under partial information through multi-turn dialogue or agent-based environments~\citep{li2024mediq,tu2025conversational,saab2026multimodal,liu2025interactive,fan2025aihospital,sviridov2025_3mdbench,chen2026clindiag,werthaim2026q4dx,ferber2026autonomous,liang2026medsp1000}. More recent systems further study autonomous diagnostic agents that perform active information acquisition~\citep{gao2026medexagent,hsu2026medaction,sanghvi2026medxagent}. \diagtrace is complementary to this line of work by evaluating not only whether an agent reaches the diagnosis, but whether its information-request trajectory matches a source-derived reference~\citep{yue2026interactivebenchmarks}. Our prior work explored source-grounded diagnostic dialogues \citep{wang2026progressive}, while DiagTrace focuses on evaluating interactive evidence acquisition.

\section{Methods}
\label{sec:methods}

\begin{figure}[t]
\centering
\includegraphics[width=.95\columnwidth]{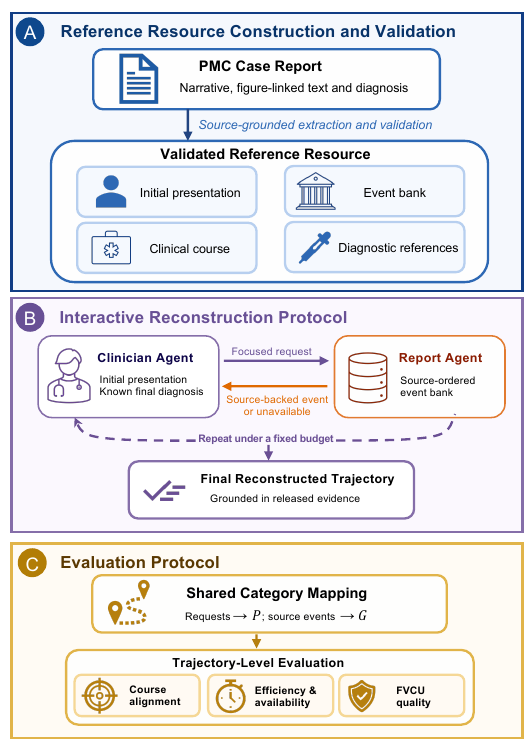}

\caption{Overview of \diagtrace. 
(A) PMC case reports are converted into validated reference resources. 
(B) A diagnosis-aware clinician agent interacts with a source-grounded report agent under a fixed turn budget to reconstruct the documented diagnostic trajectory. 
(C) Interactions are evaluated using clinical course alignment, interactive quality, turn efficiency and availability.
}
\label{fig:diagtrace-workflow}
\end{figure}

Our pipeline has three stages: (1) reference resource construction, (2) an interactive protocol, and (3) an evaluation protocol (Figure~\ref{fig:diagtrace-workflow}).

\subsection{Reference resource construction}
\label{subsec:benchmark-overview}

\textbf{Construction.}
We selected PMC case reports from eight clinical specialties (Table~\ref{tab:case-distribution}). 
Each report is converted into a structured reference resource with four components (Prompt~\ref{prompt:reference-resource}):
\begin{enumerate*}[label=(\arabic*), nosep]
\item \textit{Initial presentation}, describing the patient's presenting context; 

\item \textit{Event bank}, containing documented findings across 10 types (Table~\ref{tab:evidence-taxonomy});

\item \textit{Clinical course}, capturing the patient's progression, including diagnostic workup, treatment, follow-up, and outcomes when available; and

\item \textit{Diagnostic references}, including the final diagnosis and explicitly reported diagnostic reasoning.
\end{enumerate*}

\textbf{Data completeness validation.}
The reference resource is validated before use to ensure schema completeness, source grounding, ordering consistency, and leakage control. Source events must preserve provenance, populate required fields, and correctly mark diagnosis-revealing content for protection. The interactive experiments use separately extracted, source-quoted event banks available for 502 cases.

\textbf{Quality audit.}
We audit 100 sampled cases using GPT-4.1 mini and independently repeat the evaluation with Gemini 2.5 Flash using identical frozen inputs, prompts, and scoring procedures. Both audits use three metrics (Appendix~\ref{app:reference-resource-audit-metrics} and Prompt~\ref{prompt:reference-resource-audit}):
\begin{enumerate*}[label=(\roman*)]
\item \textit{Finding recall},
\item \textit{Event faithfulness}, and
\item \textit{Reasoning support}.
\end{enumerate*}
For comparison, we use strict scoring for finding recall and reasoning support, and lenient scoring for event faithfulness.

\textbf{Human source-event audit.}
Two independent annotators reviewed 52 unique extracted events against the original PMC articles. We sampled eight articles from the 225-case evaluation subset, one per specialty, and selected five events per article. We additionally examined the complete 11-event trajectories of two articles, yielding 52 unique events after accounting for overlap. Annotators assessed source support and fidelity without access to automated judgments, event categories, or diagnostic-role labels, and also examined the complete trajectories for source-order consistency.

\subsection{Interactive reconstruction protocol}
\label{subsec:interactive-reconstruction}

A representative episode is shown in Figure~\ref{fig:interactive-reconstruction-example}. 
Each benchmark instance initializes two agents. The clinician agent is the evaluated agent and receives the initial presentation and the final diagnosis. The report agent is initialized separately and grounded in the reference resource. All interactions are text-only; image-related evidence is conveyed through source-derived text and figure captions rather than raw image pixels.

At each interactive turn, the clinician agent issues a focused request, such as asking for additional history, examination findings, test results, or clinical-course information. Request types follow prior medical-agent and clinical-interaction benchmarks~\citep{ferber2026autonomous,li2024mediq} and are summarized in Table~\ref{tab:evidence-taxonomy}. The report agent returns at most one matching source-backed event from the reference resource, or an \emph{unavailable} status. When multiple events match, the earliest unreleased eligible event in the clinical course is returned. The report agent cannot interpret findings, infer undocumented content, or reveal the final diagnosis.

The interaction terminates when the clinician agent submits a reconstructed diagnostic trajectory or exhausts the turn budget. The resulting trace records the clinician's requests, report-agent responses, and the final reconstructed course.

The clinician-agent and report-agent prompts are provided in Prompt~\ref{prompt:interactive-reconstruction}.

\subsection{Evaluation protocol}
\label{subsubsec:evaluation-protocol}
We evaluate the interaction at the trajectory level. Both clinician requests and response events are mapped to a shared clinical information-category taxonomy (Table~\ref{tab:evidence-taxonomy}). This allows comparison between the predicted and reference clinical course even when wording differs. 

Since not all categories contribute equally to diagnostic reconstruction, we heuristically assign each category a weight reflecting its relative importance. These weights are used consistently across all weighted evaluation metrics.
Formal definitions are provided in Appendix~\ref{app:evaluation metrics}.

\textbf{Clinical Course Alignment.}
We report three metrics.
\begin{enumerate*}[label=(\arabic*)]
\item \textit{Tversky alignment} measuring category-level coverage with penalties for missed and unnecessary requests.
\item \textit{Longest-common-subsequence recall} measuring recovery of the source clinical course in relative order. 
\item \textit{Pairwise order accuracy} measuring whether two request pairs preserve source order. 
\end{enumerate*}

\textbf{Coverage and request behavior.}
To quantify whether the clinician agent requests the right clinical information categories compared with the source-derived evidence categories, we report precision, recall, and F1 using the request categories that are matched, missed, and additionally requested, respectively.
We also report the \textit{Off-protocol rate} and \textit{Redundant rate} to characterize invalid or repeated information acquisition.

\textbf{Event-level recall.} To complement category-level metrics, we measure the fraction of eligible source events released during each interaction, excluding events already provided in the initial presentation. Event-level recall is computed separately for each case and then macro-averaged across the evaluation cases. Unlike category-level recall, this metric captures the retrieval of multiple distinct events within the same information category.

\textbf{LLM-as-judge.}
We use LLM-as-judge to evaluate four 5-point dimensions (Prompt~\ref{prompt:fvcu}): 
\begin{enumerate*}[label=(\arabic*)]
\item \textit{Factuality}, 
\item \textit{Validity}, 
\item \textit{Coherence}, and 
\item \textit{Utility}. 
\end{enumerate*}

\textbf{Turn efficiency and availability.}
We report the \textit{mean turns used} under the fixed 15-turn budget. We also track \textit{answer yield}, the fraction of intermediate turns in which the report agent returns a source-backed event, and \textit{unavailable-turn rate}, the fraction of intermediate turns in which no source-backed event is returned. 

\textbf{Fixed-checklist baseline.}
To examine whether category-level metrics reward non-adaptive information seeking, we evaluate an untuned fixed-checklist policy on 16 specialty-stratified cases from the matched subset. The policy issues the same sequence of clinical requests for every case without using the diagnosis, case content, or previous responses. It interacts with the unchanged report agent under the same 15-turn budget. For comparison, we truncate the checklist trace to each model's case-specific number of acquisition turns and compute paired differences in macro-averaged event-level recall (Appendix~\ref{app:fixed-checklist}).

\paragraph{Full-scale sensitivity.} To assess the robustness of the matched-subset results, we additionally compute trajectory-alignment, weighted F1, and event-level recall metrics on each model's full set of completed cases (Appendix~\ref{app:full-scale-sensitivity}).

\subsection{Experimental setup}
\label{sec:experimental-settings}

We evaluate four clinician agents: GPT-4.1 mini~\citep{openai2025gpt41mini}, Claude Haiku 4.5~\citep{anthropic2025claudehaiku45}, Gemini 2.5 Flash~\citep{google2026gemini25flash}, and Kimi K2.5~\citep{moonshot2026kimik25} (Appendix \ref{app:implementation}). Each interaction is run once per case with a maximum budget of 15 clinician turns. GPT-4.1 mini, Claude Haiku 4.5, Gemini 2.5 Flash, and Kimi K2.5 completed 502, 498, 502, and 225 cases, respectively. We report matched comparisons on the 225-case intersection, with Kimi's lower completion count due to API failures and throughput limits rather than performance-based selection.

GPT-4.1 mini is used for benchmark construction, benchmark-quality audits, and FVCU quality evaluation. It also serves as the report agent during interactive reconstruction.

\section{Results}
\label{sec:results}

\subsection{Statistics of \diagtrace}

\diagtrace contains 510 selected PMC case reports across eight specialties (Table~\ref{tab:case-distribution}). Source-grounded construction succeeded for all selected cases, producing 5,911 source events. Each case contains an average of 11.59 source events. The event bank is clinically diverse: the most frequent evidence types are imaging, history, routine laboratory tests, treatment response, physical examination, pathology, procedures, and outcomes (Table~\ref{tab:evidence-taxonomy}). A substantial fraction of events are linked to article figures or captions, reflecting the image-rich nature of PMC case reports.
\begin{table}[t]
\centering
\small
\begin{tabular}{lrrr}
\toprule
\textbf{Specialty} & \textbf{Cases} & \textbf{Events} & \makecell[r]{\textbf{Figure-linked}\\\textbf{rate}} \\
\midrule
Dermatology        & 49  & 541  & 0.325 \\
Emergency Medicine & 25  & 307  & 0.218 \\
Internal Medicine  & 24  & 293  & 0.270 \\
Nephrology         & 29  & 366  & 0.235 \\
Neurology          & 95  & 1,128 & 0.239 \\
Oncology           & 112 & 1,284 & 0.321 \\
Ophthalmology      & 43  & 518  & 0.311 \\
Radiology          & 133 & 1,474 & 0.325 \\
~~~~\textbf{All} & \textbf{510} & \textbf{5,911} & \textbf{0.293} \\
\bottomrule
\end{tabular}
\caption{Characteristics of \diagtrace. 
}
\label{tab:case-distribution}
\end{table}

\begin{table}[h!]
\centering
\small
\begin{tabular}{lrr}
\toprule
\textbf{Category} & \textbf{Weight} & \textbf{Distribution}\\
\midrule
Pathology               & 2.00 & 7.2\% \\
Imaging                 & 1.50 & 22.7\% \\
Laboratory tests        & 1.00 & 14.9\% \\
Microbiology            & 1.00 & 2.7\% \\
Procedure/intervention  & 1.00 & 5.8\% \\
Treatment/medication    & 0.75 & 10.8\% \\
Admission/course        & 0.75 & 6.7\% \\
Clinical history        & 0.50 & 13.3\% \\
Physical examination    & 0.50 & 8.2\% \\
Other/off-protocol      & 0.25 & 7.7\% \\
\bottomrule
\end{tabular}
\caption{Request/response taxonomy
}
\label{tab:evidence-taxonomy}
\end{table}

\subsection{Reference resource quality}
\label{subsec:reference-resource-audit-results}

\textbf{LLM-based quality audit.} Because \diagtrace relies on LLM-assisted construction, we audited 100 sampled cases using GPT-4.1 mini and independently repeated the evaluation with Gemini 2.5 Flash. GPT-4.1 mini achieved 0.982 finding recall, 0.999 event faithfulness, and 0.988 reasoning support, while Gemini obtained 0.955, 0.997, and 0.992, respectively. These scores use strict criteria for finding recall and reasoning support and lenient criteria for event faithfulness. The two audits agreed on 97.61\% of event labels (1,143/1,171) and 98.30\% of reasoning labels (520/529). The similar aggregate scores suggest that the quality estimates are not specific to GPT-4.1 mini, although both audits remain LLM-based.

\textbf{Human source-event audit.} Two independent annotators reviewed 52 unique extracted events against the original PMC articles and agreed on 48 labels (92.3\%; Cohen's $\kappa=0.874$; weighted $\kappa=0.925$). The annotators judged 50/52 and 52/52 events, respectively, to be supported by the source articles. For 20 events overlapping with the original GPT-4.1 mini audit, GPT's faithfulness judgments agreed with the two annotators on 18/20 and 20/20 events, respectively. Both annotators identified discrepancies in article narrative order in the two fully audited trajectories (Appendix~\ref{app:supplementary-audit-findings}). These findings support the source grounding of the sampled events but do not establish complete finding coverage or source-order accuracy.

\subsection{Interactive reconstruction results}
\label{subsec:interactive-reconstruction-results}

Table \ref{tab:interactive-model-comparison} reports interactive reconstruction results on the 225-case subset.
The models show different trade-offs between category coverage, source-consistent ordering, and interaction efficiency. Kimi K2.5 achieves the highest Tversky alignment (0.399) and LCS recall (0.561), while Claude Haiku 4.5 achieves the highest pairwise order accuracy (0.690). GPT-4.1 mini and Gemini 2.5 Flash show comparable Tversky alignment (0.360 and 0.359, respectively). GPT-4.1 mini uses the fewest turns on average, whereas Gemini 2.5 Flash uses the most turns and has a higher unavailable-turn rate.

\begin{table}[t]
\centering
\small
\setlength{\tabcolsep}{4pt}

\begin{tabular}{lcccc}
\toprule
\textbf{Metric}
& \textbf{GPT}
& \textbf{Claude}
& \textbf{Gemini}
& \textbf{Kimi} \\
\midrule

\rowcolor{gray!20}
\multicolumn{5}{l}{\textbf{Request-category alignment}} \\
Tversky alignment
& 0.360 & 0.334 & 0.359 & 0.399 \\
LCS recall
& 0.465 & 0.510 & 0.500 & 0.561 \\
Pairwise order accuracy
& 0.646 & 0.690 & 0.654 & 0.601 \\

\rowcolor{gray!20}
\multicolumn{5}{l}{\textbf{Coverage decomposition and request behavior}} \\
Precision
& 0.626 & 0.560 & 0.607 & 0.623 \\
Recall
& 0.525 & 0.582 & 0.573 & 0.672 \\
F1
& 0.592 & 0.589 & 0.603 & 0.638 \\
Event-level recall
& 0.577 & 0.665 & 0.655 & 0.661 \\
Off-protocol rate
& 0.000 & 0.003 & 0.005 & 0.002 \\
Redundant rate
& 0.099 & 0.170 & 0.150 & 0.202 \\

\rowcolor{gray!20}
\multicolumn{5}{l}{\textbf{Interactive FVCU quality}} \\
Factuality
& 4.81 & 4.82 & 4.75 & 4.85 \\
Validity
& 4.84 & 4.84 & 4.78 & 4.87 \\
Coherence
& 4.84 & 4.87 & 4.79 & 4.92 \\
Utility
& 4.22 & 4.20 & 4.11 & 4.29 \\

\rowcolor{gray!20}
\multicolumn{5}{l}{\textbf{Turn efficiency and availability}} \\
Mean turns used
& 9.93 & 11.31 & 11.88 & 10.28 \\
Answer yield
& 0.525 & 0.611 & 0.544 & 0.642 \\
Unavailable-turn rate
& 0.351 & 0.272 & 0.348 & 0.250 \\

\bottomrule
\end{tabular}

\caption{
Interactive reconstruction results.
}
\label{tab:interactive-model-comparison}
\end{table}

\paragraph{Coverage decomposition and request behavior.}
Kimi K2.5 achieves the highest weighted recall (0.672) and F1 (0.638), while GPT-4.1 mini obtains the highest precision (0.626). GPT-4.1 mini also has the lowest redundant category rate, reflecting a more concise information-seeking strategy. Off-protocol request rates remain low across all models, suggesting that differences mainly arise from category selection, omission, ordering, and repeated requests rather than invalid interaction behavior.

\paragraph{Event-level recall.}
Claude Haiku 4.5 achieves the highest macro-averaged event-level recall (0.665), followed by Kimi K2.5 (0.661), Gemini 2.5 Flash (0.655), and GPT-4.1 mini (0.577). Although Kimi achieves the highest category-level alignment, Claude retrieves the largest proportion of eligible source events. This difference suggests that category-level alignment and event-level evidence acquisition capture complementary aspects of information-seeking behavior.

\paragraph{Fixed-checklist baseline.}
On 16 specialty-stratified cases, all four clinician models achieved higher event-level recall than the fixed checklist at matched turn budgets, with mean paired improvements of 0.022 (GPT), 0.097 (Claude), 0.159 (Gemini), and 0.207 (Kimi) (Table~\ref{tab:fixed-checklist}). The 95\% paired-bootstrap confidence intervals excluded zero for Gemini and Kimi, but not for GPT or Claude. These exploratory results suggest that adaptive information seeking can improve evidence recovery beyond a fixed request sequence.

\begin{figure}[t]
\centering
\includegraphics[
    width=\columnwidth,
    clip
]{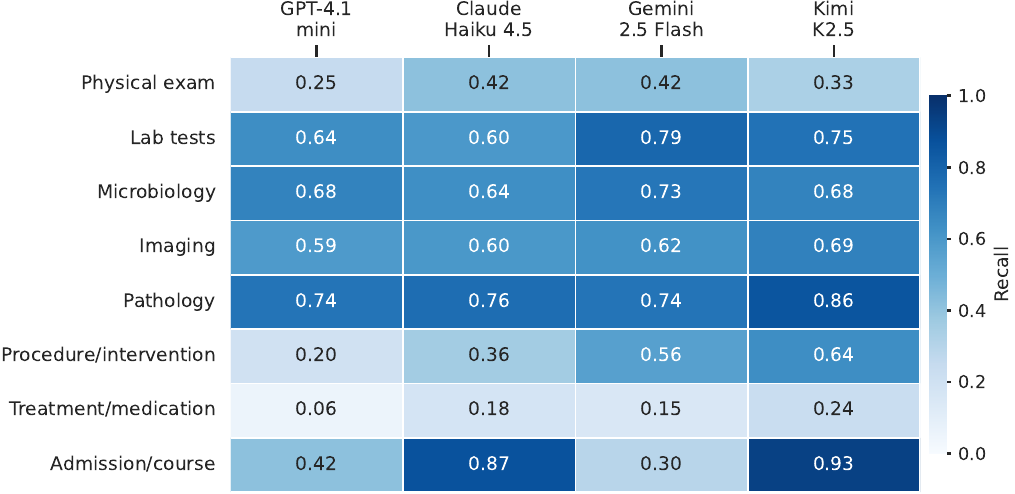}
\caption{Category-wise recall. Each cell is the fraction of cases in which a source-derived category was requested by the clinician agent.}
\label{fig:category-recall-heatmap}
\end{figure}

\paragraph{Category-wise results.} 
Laboratory, microbiology, imaging, and pathology evidence is requested relatively consistently across models, whereas treatment-related and physical-examination evidence is requested less frequently. Procedure/intervention and admission/course categories show greater variation across models (Figure~\ref{fig:category-recall-heatmap}).

\paragraph{Full-scale sensitivity.} Results on the full completed-case sets (502 GPT, 498 Claude, and 502 Gemini cases) were broadly consistent with the matched-225 results, with absolute changes of at most 0.037 across five principal metrics (Table~\ref{tab:full-scale-sensitivity}). Full-scale estimates are descriptive because the case sets differ across models.

Overall, the results suggest that stronger category coverage does not always coincide with the best pairwise ordering or the shortest interaction.

\section{Conclusion}

We introduced \diagtrace, a source-grounded benchmark for evaluating interactive reconstruction of diagnostic trajectory from clinical case reports. Unlike static diagnosis benchmarks, it measures how models acquire supporting information. Experiments on 225 cases reveal trade-offs in coverage, ordering, and interaction efficiency.

\clearpage

\section*{Limitations}

The benchmark also has several limitations. First, source order reflects the published article narrative rather than an adjudicated clinical timeline. Human auditing also identified article-order discrepancies that may affect order-sensitive metrics, including LCS recall and pairwise order accuracy. Second, clinical case reports are selective summaries and may not reflect routine workflows or the full space of diagnostic interactions. Third, benchmark construction and quality auditing rely in part on LLM assistance, making continued validation important.

\section*{Ethical Considerations}

\diagtrace is intended solely for research evaluation and not for clinical decision support or patient care.
It is constructed from publicly available PMC case reports and may inherit their selection and reporting biases.
Because the clinician agent is given the final diagnosis, benchmark scores measure diagnostic-trajectory reconstruction rather than autonomous diagnostic ability and should not be interpreted as evidence of clinical safety.
LLM-assisted benchmark construction and evaluation may also introduce residual errors despite the validation procedures.

\section*{Acknowledgments}

We thank Ranxuan Li and Bicheng Chen from Weill Cornell Medicine for their contributions to the human validation of the benchmark, including the independent audit of source--event mappings. We also thank the anonymous reviewers and the area chair for their constructive feedback. This research was supported in part by the National Institutes of Health (NIH) under grants R01LM014306, R01LM014573, and
UG3TR005954.

\bibliography{custom1}

\appendix

\section{Appendix}

\subsection{Reference Resource Audit Metrics}
\label{app:reference-resource-audit-metrics}

We audit a sampled subset of benchmark references to assess source grounding, leakage control, and clinical usability.

\paragraph{Finding recall}
measures whether the event bank captures the major diagnosis-relevant findings. Under strict scoring, it is the fraction of diagnosis-relevant findings in the source case report that are judged fully represented in the extracted event bank:
$
\frac{|\mathcal{F}_{\mathrm{full}}|}
{|\mathcal{F}|}.
$

\paragraph{Event faithfulness}
assesses whether extracted events are supported by source quotes. Under lenient scoring, it is the fraction of extracted events that are judged faithful or partially faithful to their supporting source quotations:
$
\frac{|\mathcal{E}_{\mathrm{faithful}}|+|\mathcal{E}_{\mathrm{partial}}|}
{|\mathcal{E}|}.
$

\paragraph{Reasoning support}
evaluates whether the event bank contains information supporting the article-derived diagnostic rationale. Under strict scoring, it is the fraction of article-derived diagnostic reasoning items that are judged fully supported by the extracted event bank:
$
\frac{|\mathcal{R}_{\mathrm{full}}|}
{|\mathcal{R}_{\mathrm{ref}}|}.
$

\subsection{Supplementary Audit Findings}
\label{app:supplementary-audit-findings}

\paragraph{Article-order discrepancies.}
Both annotators identified article-order discrepancies in the two fully reviewed trajectories (PMC10906755 and PMC9721097). Although the extracted events were source-supported, their ordering differed from the published narratives. These discrepancies concern article presentation order rather than verified clinical chronology and may affect order-sensitive evaluation metrics.

\subsection{Fixed-Checklist Baseline}
\label{app:fixed-checklist}

\paragraph{Sampling and protocol.}
We sampled 16 cases from the shared 225-case subset, with two cases per specialty, using a fixed random seed (20260916). The checklist followed 14 acquisition requests in a fixed order: history, physical examination, laboratory tests, microbiology, imaging, pathology, procedures, treatment response, clinical course, and then repeated history, examination, laboratory tests, microbiology, and imaging. The final turn was reserved for trajectory submission. The policy did not access case-specific information or previous responses and used the unchanged GPT-4.1 mini report agent. All 16 checklist episodes completed successfully.

\paragraph{Turn matching and evaluation.}
To account for differences in interaction length, each checklist trajectory was truncated to the corresponding clinician model's number of non-final turns for that case. Event-level recall was calculated using unique event IDs returned with a \texttt{found} status, divided by the number of eligible reference events not included in the initial presentation. We calculated recall separately for each case and averaged across the 16 cases. Model-minus-checklist differences were computed for each paired case, with 95\% confidence intervals estimated using 10,000 paired bootstrap resamples.

\begin{table}[H]
\centering
\small
\resizebox{\columnwidth}{!}{
\begin{tabular}{lcccc}
\toprule
\textbf{Model} & \textbf{Model} & \textbf{Checklist} & $\boldsymbol{\Delta}$ & \textbf{95\% CI} \\
\midrule
GPT & 0.546 & 0.524 & +0.022 & $[-0.055, 0.104]$ \\
Claude & 0.638 & 0.540 & +0.097 & $[-0.019, 0.216]$ \\
Gemini & 0.689 & 0.531 & +0.159 & $[0.047, 0.277]$ \\
Kimi & 0.656 & 0.449 & +0.207 & $[0.108, 0.327]$ \\
\bottomrule
\end{tabular}
}
\caption{Event-level recall on the 16-case fixed-checklist evaluation. Checklist scores are matched to each model's case-specific acquisition-turn budget. Differences and confidence intervals are based on paired case-level comparisons.}
\label{tab:fixed-checklist}
\end{table}

\subsection{Full-scale Sensitivity Analysis}
\label{app:full-scale-sensitivity}

We recomputed the evaluation metrics from saved completed episodes using the corrected request-category classifier and canonical specialty--PMCID case identities. Table~\ref{tab:full-scale-sensitivity} compares results on the shared 225-case subset with each model's full set of completed cases.

\begin{table}[t]
\centering
\scriptsize
\setlength{\tabcolsep}{3pt}
\begin{tabular}{lccccc}
\toprule
Model ($n$) & Tversky & LCS & Pair. & F1 & Event R \\
\midrule
GPT (225) & .360 & .465 & .646 & .592 & .577 \\
GPT (502) & .393 & .502 & .669 & .618 & .575 \\
\midrule
Claude (225) & .334 & .510 & .690 & .589 & .665 \\
Claude (498) & .362 & .537 & .679 & .610 & .666 \\
\midrule
Gemini (225) & .359 & .500 & .654 & .603 & .655 \\
Gemini (502) & .380 & .535 & .670 & .622 & .654 \\
\midrule
Kimi (225) & .399 & .561 & .601 & .638 & .661 \\
\bottomrule
\end{tabular}
\caption{Matched-subset and full completed-case sensitivity results. Pair. denotes pairwise order accuracy; Event R denotes event-level recall. Full-case estimates are not strictly paired across models.}
\label{tab:full-scale-sensitivity}
\end{table}

Across GPT-4.1 mini, Claude Haiku 4.5, and Gemini 2.5 Flash, absolute changes in the five principal metrics were at most 0.037. Kimi K2.5 had no additional completed cases beyond the matched subset. The matched subset was more balanced across specialties, whereas the remaining completed cases were concentrated in radiology and oncology. Thus, the stability of aggregate results does not establish that the matched subset is representative of the full dataset.

\subsection{Evaluation metrics}
\label{app:evaluation metrics}

\subsubsection{Clinical Course Alignment Metrics}
\label{app:request-category-metrics}

The interaction trace is converted into a clinical-request sequence $P$ and a reference sequence $G$. Both requests and responses are mapped to the shared taxonomy (Table~\ref{tab:evidence-taxonomy}). 
Overly broad requests or requests for retrospective diagnostic justification are mapped to \texttt{Other/off-protocol}. 

To emphasize category coverage and clinical course order, adjacent duplicate categories are first collapsed, and only the first occurrence of each category is retained. The resulting compact sequences therefore contain at most one occurrence of each request category.

The initial presentation is removed because it is provided before the interaction begins. 
Final-turn request/response and system-control turns (e.g., generic wait or reassessment without a concrete information request/response) are also excluded. 

Reassessment turns that explicitly request documented clinical-course updates, follow-up status, interval change, or treatment response are retained.
Requests that receive \texttt{unavailable} responses are also retained, because they reflect the clinician agent's information-seeking strategy under a fixed turn budget. 

Evaluation compares the generated sequence $P$ with the reference sequence $G$ using the following metrics.

\paragraph{Tversky alignment} measures category coverage while penalizing both missed and unnecessary requests.
Let $W_{\mathrm{match}}$, $W_{\mathrm{missed}}$, and $W_{\mathrm{extra}}$ denote the total weights of categories that appear in both $G$ and $P$, only in $G$, and only in $P$, respectively. The score is defined as:
\begin{equation}
T_{\mathrm{w}}
=
\frac{W_{\mathrm{match}}}
{W_{\mathrm{match}}+\alpha W_{\mathrm{extra}}+\beta W_{\mathrm{missed}}}.
\end{equation}
Following \cite{ferber2026autonomous}, we set $\alpha=2$ and $\beta=1$, placing a stronger penalty on unnecessary requests because they consume the limited turn budget.

\paragraph{Longest-common-subsequence (LCS) recall} measures the extent to which $G$ is recovered in the correct relative order:
\begin{equation}
R_{\mathrm{wLCS}}
=
\frac{
\sum_{c \in \operatorname{LCS}(G,P)} w(c)
}{
\sum_{c \in G} w(c)
}.
\end{equation}
Because this metric is recall-oriented, additional requests do not directly reduce the score, but missing requests and order changes can reduce the score. 

\paragraph{Pairwise order accuracy}
evaluates ordering consistency among shared categories. Let $\mathcal{C}(G,P)$ be the set of category pairs $(a,b)$ such that $a$ precedes $b$ in $G$ and both categories also occur in $P$. Each pair receives weight:
\begin{equation}
\omega(a,b)=\sqrt{w(a)w(b)}.
\end{equation}
Let indicator $\mathbbm{1}_P(a,b)$ denote if $a$ precedes $b$ in $P$. The score is defined as:
\begin{equation}
A_{\mathrm{pair}}
=
\frac{
\sum_{(a,b)\in\mathcal{C}(G,P)}
\omega(a,b) \mathbbm{1}_P(a,b)
}{
\sum_{(a,b)\in\mathcal{C}(G,P)}
\omega(a,b)
}.
\end{equation}
Cases with fewer than two shared categories are excluded.

\subsubsection{Coverage and request behavior metrics}
\label{app:interactive-extra-metrics}

For each case, let $W_{\mathrm{match}}$, $W_{\mathrm{missed}}$, and $W_{\mathrm{extra}}$ denote the total weights of request categories that are matched, missed, and additionally requested, respectively, as defined in the weighted Tversky alignment.
The metrics are defined as:
\begin{align}
P_{\mathrm{w}}
&=
\frac{W_{\mathrm{match}}}
{W_{\mathrm{match}} + W_{\mathrm{extra}}},
\\
R_{\mathrm{w}}
&=
\frac{W_{\mathrm{match}}}
{W_{\mathrm{match}} + W_{\mathrm{missed}}},\\
F_{1,\mathrm{w}}
& =
\frac{2P_{\mathrm{w}}R_{\mathrm{w}}}
{P_{\mathrm{w}} + R_{\mathrm{w}}}.
\label{eq:weighted-f1}
\end{align}

\paragraph{Off-protocol rate} measures the proportion of intermediate interaction turns that produce an off-protocol report-agent status:
\begin{equation}
R_{\mathrm{off}}
=
\frac{N_{\mathrm{off}}}
{N_{\mathrm{intermediate}}},
\end{equation}
where $N_{\mathrm{off}}$ is the number of intermediate turns containing an overly broad, invalid, or overly diagnostic request, and $N_{\mathrm{intermediate}}$ is the total number of intermediate interaction turns.

\paragraph{Redundant rate} measures whether the clinician agent re-issues a request.
Let $P_{\mathrm{raw}}$ denote the request sequence after collapsing adjacent duplicate requests.
A request is considered redundant when it has appeared previously in $P_{\mathrm{raw}}$.
The redundant rate is defined as:
\begin{equation}
R_{\mathrm{red}}
=
\frac{N_{\mathrm{red}}}
{|P_{\mathrm{raw}}|},
\end{equation}
where $N_{\mathrm{red}}$ is the number of redundant category requests.

\subsection{Implementation details}
\label{app:implementation}

Each interactive reconstruction used a fixed budget of 15 clinician-report turns. Each clinical agent was evaluated once per case without sampling multiple stochastic trajectories.

All API requests used a 600-second timeout. For Claude Haiku 4.5, Gemini 2.5 Flash, and Kimi K2.5, we set temperature to 0 and the maximum output length to 4,096 tokens. Top-$p$, seed, and reasoning-effort parameters were left at their provider defaults. 

GPT-4.1 mini was accessed through Azure OpenAI using JSON-mode chat-completion calls with the deployment-level decoding configuration.
Temperature, top-$p$, maximum output length, seed, and reasoning effort were not specified. 

Transient connection and timeout failures were retried using the same prompts and inference configuration.

\subsection{Data and Code Availability}
\label{app:data-code-availability}

The DiagTrace benchmark data and interaction traces are publicly available at \url{https://github.com/Fanncg1229/Diagtrace}. The corrected camera-ready release includes the canonical event banks, matched and full-scale interaction episodes, evaluation manifests, and fixed-checklist artifacts under \texttt{camera\_ready\_release/}. The evaluation code, corrected aggregate results, and reproducibility instructions are available at \url{https://github.com/Fanncg1229/DiagTrace-code}. The repositories provide instructions for reproducing the deterministic evaluation metrics and figures.

\newpage
\subsection{Prompts}
\label{app:prompts}

\begin{prompt}[Reference resource construction]
\label{prompt:reference-resource}
Given the complete PMC case report, construct a source-grounded reference resource for an interactive diagnostic benchmark.

Extract: (1) an initial presentation that does not reveal the final diagnosis; (2) a source-ordered event bank of documented clinical findings; (3) clinical-course information derived from the event bank; and (4) article-derived diagnostic references, including the final diagnosis and diagnostic reasoning when explicitly available.

Each source event must be grounded in the report text and include provenance, source order, evidence type, diagnostic role, and figure or caption linkage when applicable. Do not infer unsupported findings, introduce external medical knowledge, or treat article order as an adjudicated clinical timeline. Mark diagnosis-revealing content for protected handling. Return valid JSON only.
\end{prompt}

\begin{prompt}[Reference resource quality audit]
\label{prompt:reference-resource-audit}
You are auditing a source-grounded reference resource for an interactive medical diagnosis benchmark. Evaluate benchmark data quality, not downstream model performance. Use only the supplied source report text and extracted reference resource. Do not add external medical facts.

First, identify a compact inventory of 8--15 major diagnosis-relevant findings from the source report. The inventory should span the diagnostic trajectory when present, including initial presentation, history, examination, laboratory or microbiology results, imaging, pathology or molecular confirmation, procedures, treatment response, follow-up, and clinical course.

For each source finding, decide whether it is represented in the source-ordered event bank using one of four labels: \texttt{full}, \texttt{partial}, \texttt{none}, or \texttt{contradicted}.

For each extracted event, decide whether it is faithful to its supporting source quotation using one of four labels: \texttt{faithful}, \texttt{partial}, \texttt{unsupported}, or \texttt{contradicted}.

For each article-derived diagnostic reasoning item, decide whether it is supported by the source-ordered event bank and diagnostic references using \texttt{full}, \texttt{partial}, \texttt{none}, or \texttt{contradicted}.

Return valid JSON only.
\end{prompt}

\newpage

\begin{prompt}[Interactive reconstruction protocol]
\label{prompt:interactive-reconstruction}
\textbf{Diagnosis-aware clinician agent.}
You are given an initial presentation and the article-reported final diagnosis. Your task is not to re-diagnose the patient, but to reconstruct the source-derived diagnostic trajectory that supports the known diagnosis. At each non-final turn, issue one focused clinical request, such as asking about history, examination findings, tests, documented results, procedures, treatment response, follow-up, or clinical-course updates.

Use the known diagnosis to guide information seeking, but treat only report-agent replies as observed case information. Do not treat textbook expectations or general medical knowledge as facts about the case. Maintain a working trajectory reconstruction, confirmed findings, remaining information gaps, and confidence in the sufficiency of the acquired information. When sufficient information has been acquired, submit a final reconstructed trajectory grounded only in released event identifiers. Return valid JSON only.

\textbf{Source-grounded report agent.}
You are given the source-ordered event bank extracted from the case report. For each clinician request, return at most one responsive, previously unreleased, source-backed event, or return an unavailable status if no eligible event is documented. When multiple events match, prefer the earliest eligible event in preserved source order.

Do not diagnose, interpret clinical significance, recommend next steps, infer undocumented facts, or expose retrospective diagnostic synthesis. Do not convert absence from the source into a negative clinical finding. Return valid JSON only.
\end{prompt}

\newpage

\begin{prompt}[Interactive FVCU quality audit]
\label{prompt:fvcu}
You are evaluating an interactive diagnostic-trajectory reconstruction episode. Evaluate the diagnosis-aware clinician agent, not the report agent. Use only the provided reference resource and interaction trace.

Score four FVCU dimensions on a 1--5 scale.

\textbf{Factuality}: whether the clinician agent's case-specific claims, confirmed findings, and final reconstruction are grounded in report-agent responses available by that turn.

\textbf{Validity}: whether the clinician agent's requests and interpretations are clinically reasonable given the known diagnosis and acquired information.

\textbf{Coherence}: whether the working reconstruction, confirmed findings, missing information, and confidence evolve consistently across turns.

\textbf{Utility}: whether the clinician agent uses the fixed turn budget to acquire high-yield documented information needed to reconstruct the source-derived diagnostic trajectory.

Return valid JSON only, including one score and brief rationale for each dimension.
\end{prompt}

\begin{figure}[t]
\centering
\includegraphics[width=\linewidth]{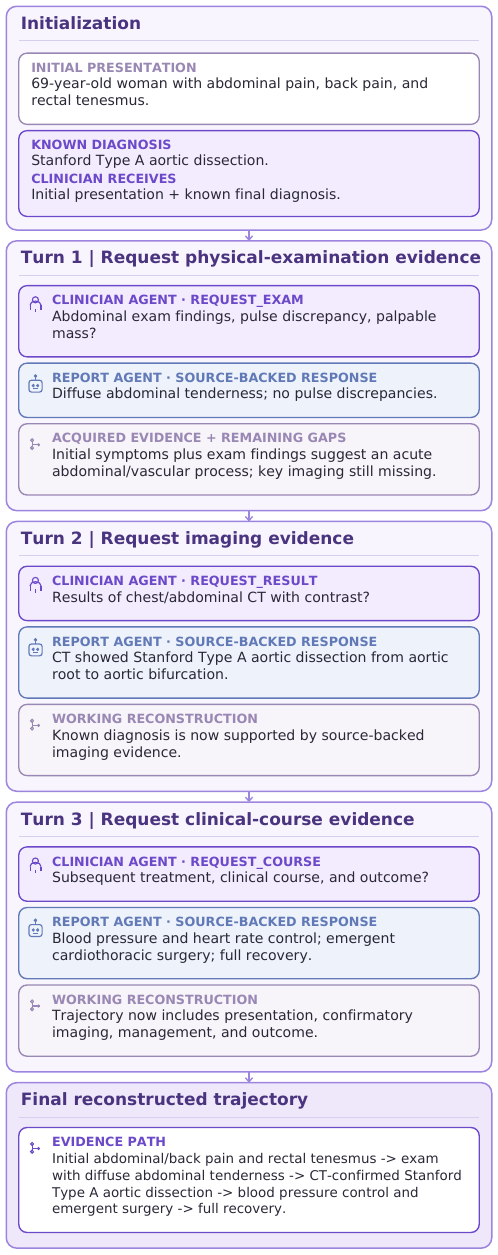}
\caption{Example diagnosis-aware interactive reconstruction episode. The clinician agent is initialized with the initial presentation and known diagnosis, actively requests source-backed information, and reconstructs the diagnostic trajectory under a fixed turn budget.}
\label{fig:interactive-reconstruction-example}
\end{figure}

\end{document}